\documentclass[lettersize,journal]{IEEEtran}
\usepackage{amsmath,amsfonts}
\usepackage{algorithmic}
\usepackage{algorithm}
\usepackage{array}
\usepackage[caption=false,font=normalsize,labelfont=sf,textfont=sf]{subfig}
\usepackage{textcomp}
\usepackage{stfloats}
\usepackage{url}
\usepackage{verbatim}
\usepackage{graphicx}
\usepackage{cite}
\begin{document}

\title{Identifying Information-Transfer Nodes in a Recurrent Neural Network Reveals Dynamic Representations}

\author{Arend Hintze, Asadullah Najam, Jory Schossau}



\maketitle

\begin{abstract}
Understanding the internal dynamics of Recurrent Neural Networks (RNNs) is crucial for advancing their interpretability and improving their design. This study introduces an innovative information-theoretic method to identify and analyze information-transfer nodes within RNNs, which we refer to as \textit{information relays}. By quantifying the mutual information between input and output vectors across nodes, our approach pinpoints critical pathways through which information flows during network operations. We apply this methodology to both synthetic and real-world time series classification tasks, employing various RNN architectures, including Long Short-Term Memory (LSTM) networks and Gated Recurrent Units (GRUs). Our results reveal distinct patterns of information relay across different architectures, offering insights into how information is processed and maintained over time. Additionally, we conduct node knockout experiments to assess the functional importance of identified nodes, significantly contributing to explainable artificial intelligence by elucidating how specific nodes influence overall network behavior. This study not only enhances our understanding of the complex mechanisms driving RNNs but also provides a valuable tool for designing more robust and interpretable neural networks.
\end{abstract}


\section{Introduction}
\IEEEPARstart{A}{rtificial} neural networks represent a significant enigma. On the one hand, we grasp the mechanism of backpropagation and the dataset that optimally trains them; on the other hand, we do not fully comprehend how ANNs or their recurrent counterparts, RNNs, accomplish their tasks. At their most basic, their function can be described as follows: input vectors are propagated forward through multiple layers by applying a dot product between matrices and these vectors, alongside the application of mathematical threshold functions to the vectors. While this functional description enables us to implement and compute neural networks relatively easily, it applies to a network executing the MNIST handwritten numeral task or the CIFAR-10 image classification problem. The functional differences between these tasks lie in the variations between their weight matrices, which yield complex hidden state vectors that contain and propagate the processed information. Numerous efforts have been made to decode the roles of these vector components. Notably, these state vectors are analogous to natural neural networks firing patterns, which are similarly challenging to interpret. Regrettably, it is rare for a single neuron or node in an ANN to be directly correlated with a specific function performed by the network. For instance, in the MNIST example, a tight correlation between the state of the first node in the hidden layer and the numbers being classified is seldom observed; if it were, assigning functions to each node would not be such a challenge.

Interestingly, the endeavor to decipher the hidden state space is analogous to the longstanding notion that the states in recurrent neural networks converge on an attractor \cite{hopfield1982neural}. Hopfield himself posited that information is stored within the \textit{gestalt} of these hidden state-space attractors. This concept has spurred numerous efforts to visualize and interpret the hidden state spaces in deeply learned neural networks \cite{de2003visual,minh2022explainable}. These approaches utilize the entire state vector, whether it be the recurrent layer, the bottleneck of a variational autoencoder \cite{kingma2013auto}, a generative model \cite{goodfellow2014generative}, or simply a hidden layer. This vector is regarded as representing a single point in a phase space, and efforts can be made to segment this phase space into attractors or clusters. Through such mappings, we attempt to describe the function \footnote{This approach is so prevalent that we cannot attribute it to a single reference where it first appeared. However, we find many recent publications that seek to understand representations within the phase space \cite{horoi2020low,rajalingham2022recurrent,farrell2022gradient}}.

To further simplify, the phase space can be reduced to a 2D plot using techniques such as principal component analysis (PCA) or other projection methods, t-SNE being one of the most successful \cite{van2008visualizing}. This often results in neatly organized clusters, each correlating with, for example, one of the input classes. Such behavioral mapping of phase space is invaluable for understanding the overall function of a neural network in terms of node recruitment, depending on the input. However, this method does not disclose the function of each node in the context of the task -- or ``What does each node do?''

We have previously addressed this question by introducing the concept of information-theoretic representations \cite{marstaller2013evolution}, which pertain to the information that neural networks possess or process about their environment. This concept significantly diverges from the traditional view of representations in neural networks, which typically focuses on how networks transform and abstract input data through their layers. In contrast, information-theoretic representations emphasize the quantifiable information exchange between the network's hidden states and its external environment, independent of the current inputs. This method of quantifying representations using information theory can be readily extended to work on arbitrary computational models \cite{hintze2018structure,kirkpatrick2020evolution}, can reveal functional components \cite{bohm2022understanding}, and enhance understanding of how distributed representations improve their functionality and robustness \cite{kirkpatrick2019role,sardar2023robustness}. Recently, we expanded the concept of representations to identify what we call \textit{information relays} \cite{hintze2023detecting}. In this extension, nodes in a hidden layer that are involved in relaying information from the inputs to the outputs while performing a specific classification task can now be identified.

The method is predicated on the notion that each layer in an artificial neural network functions as an information-theoretic channel. To put it simply, each layer, disregarding the threshold function, follows the general formula $O = I \cdot W$, where $I$ represents the input vector, $O$ the output vector, and $W$ the weight matrix. The information relayed is then quantified as the mutual information between $I$ and $O$. When multiple hidden layers are involved, the information propagates from one layer to the next, yet the total information exchanged between the inputs and outputs remains as the mutual information. To determine which nodes within a hidden layer relay this information, one can calculate the information ($I_R$) contained within a subset of nodes in that hidden layer ($Y_R$), which is shared with the inputs ($X_{\rm in}$) and outputs ($X_{\rm out}$), and which is not found in the remaining set of hidden nodes ($Y_0$):

\begin{equation}
    I_R=H(X_{\rm in};X_{\rm out};Y_R|Y_0)\; \label{equ:relay}
\end{equation}

This method alone quantifies only how much information \textit{flows} through a given set of hidden nodes. However, we can also establish an ordering of the nodes such that they exhibit increasing degrees of information relaying function. The algorithm begins with the complete set and sequentially removes nodes that carry the least amount of information. This is a ``greedy algorithm'' in that it minimizes information removal by considering only the current configuration, yet it has been verified to approximate optimal node-ordering effectively. For a detailed derivation and verification of Equation \ref{equ:relay} and the greedy algorithm, see Hintze and Adami 2023.

So far, this method has only been demonstrated to be effective on artificial neural networks without recurrence. In a recurrent neural network, or any of its more sophisticated derivatives like LSTM \cite{hochreiter1997long} or GRU \cite{cho2014learning}, information is not solely propagated through the network. Instead, recurrent hidden states (or cell states in the case of an LSTM) store previous information and, together with new inputs, allow for more complex computations. Thus, recurrent networks are not merely information-theoretic channels but integrate information from previous time steps. Consider a task for a recurrent network where inputs can be -1, 0, or 1, and the recurring states of the network need to keep track of the current sum of those inputs. Now, the hidden or recurring states are not just part of the input-output channel but also contain information about the entire history of all inputs. In such a context, the information relay method would not be applicable. However, in the case of time series classification, the information of the recurrent layer might again become part of the information relay between inputs and outputs. Imagine the same example network as before, but where at the end of the sequence, the network needs to report whether the total sum was positive or negative. In this scenario, the information relay method should be able to identify the relevant recurrent states. Here, we will demonstrate that the information relay method can indeed detect the pertinent recurrent nodes in a time series classification task.

Identifying what nodes do is crucial for at least two reasons. The first is the quest for explainable artificial intelligence \cite{ribeiro2016should,samek2017explainable}, which focuses on identifying which parts of the data or process contribute to the outcome \cite{guidotti2018survey}. Our work complements this endeavor by seeking to understand the function of individual components. The second reason stems from the historical development of recurrent neural networks. Following their inception \cite{rumelhart1986learning}, it was quickly discovered that RNNs struggle with input sequences that contain long-range dependencies \cite{hochreiter1991untersuchungen}, a problem attributed to vanishing or exploding gradients \cite{hochreiter1991untersuchungen,bengio1994learning}. This weakness can be largely mitigated by altering the network's architecture. Both alternative models, LSTMs and GRUs, incorporate dedicated pathways to set and clear hidden states, leading to their tremendous success. However, these advantages seem inadequate for processing text in large language models, where the transformer architecture with its attention mechanism has made a significant difference \cite{vaswani2017attention}. Interestingly, Vaswani et al. (2017) argue that transformers allow better parallelization of training due to their non-recurrent architecture, not that LSTMs or GRUs have a principle shortcoming like their RNN ancestor did in the form of vanishing gradients. The ambiguity surrounding what \textit{self-attention} means in the context of an LSTM or GRU, which is foundational to the transformer model, remains. Nonetheless, the absence of an explanation for why LSTMs or GRUs struggle with tasks where transformer models excel invites further investigation into how nodes represent information. Therefore, after demonstrating that the information relay method can be applied to recurrent states, we will compare RNNs to LSTMs and GRUs, and discuss their differences.

Various types of recurrent neural networks were trained on time series classification tasks to demonstrate that the information relay method can pinpoint the relevant hidden nodes. These tasks are designed such that information pertaining to different aspects is fed into the network at specific time points. This information must be stored across multiple time points with varying intervals and subsequently reported. For instance, the network may be shown a moving block and later required to determine whether the block was bright or dark, small or large, and whether it was moving left or right. After applying the information relay method and identifying relevant nodes, these nodes and randomly selected nodes for control will be subjected to noise to test the network's classification ability anew. We will demonstrate that the nodes identified by the information relay method are also most susceptible to noise, thus confirming their functional contribution. Similarly, we will illustrate how the identified nodes help condense the phase space, bridging the gap to previous analyses. Further examination will explore how information is stored and distributed within the networks and reveal that, under less favorable conditions, information is even ``moving around'' within the network.

\section{Materials and Methods}
\subsection{Time Series Classification Tasks}
\begin{figure}
\centering
\includegraphics[width=0.475\textwidth]{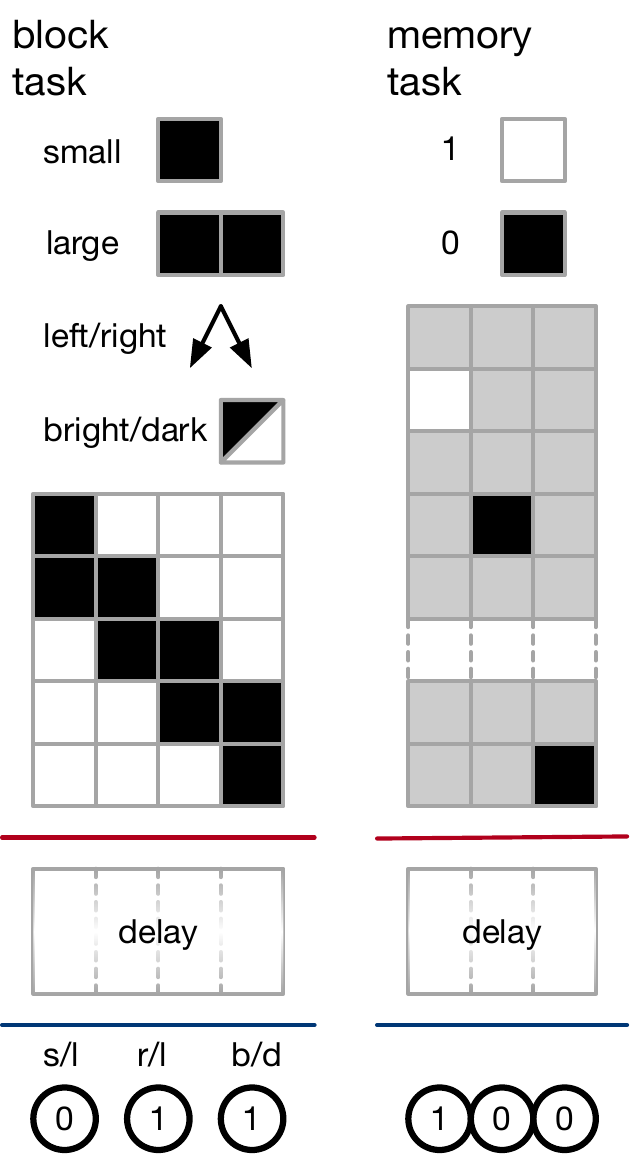}
\caption{Illustration of the block (left) and the memory (right) task.}\label{fig:taskIllustration}
\end{figure}

Two time series classification tasks were employed in this study. The first task, termed the "memory" task, is abstract and requires the neural network to monitor three sensors and subsequently report their deviation from a mean (either positive or negative). The second task involves observing a block moving before a camera, thus dubbed the "block" task. Here, the network must later report the block's size, direction, and hue. This second task is more complex as it necessitates integrating information from multiple sensors and time points to compute the outcomes to be reported later. For both tasks, a sequence of inputs is generated and fed into the neural network (RNN, LSTM, or GRU), after which the result must be reported -- not immediately, but with some delay. Based on previous experiments, we understand that the variability in these delays can profoundly impact the network's functionality \cite{hafiz2022predicting,bohm2022understanding}. Consequently, we introduce additional meaningless inputs (all input vector values are 0.0) at the end before the network must report the answer. Delays ranging from 1 to 5 time steps and a random selection from the interval $[1,5]$ are utilized. The latter approach is expected to make the networks more robust and improve temporal domain generalization. See Figure \ref{fig:taskIllustration} for an illustration of both tasks.

\subsection{RNN, GRU, and LSTM training}
RNNs, GRUs, and LSTMs each equipped with 12 recurrent nodes followed by a single linear layer that condenses those 12 states to 3 output nodes were implemented and trained using PyTorch \cite{NEURIPS2019_9015}. Each layer utilized a hyperbolic tangent threshold function. Training was conducted using the ADAM optimizer \cite{kingma2014adam} with a mean squared error loss function. The training process was terminated when an accuracy of $0.98$ was achieved or 2000 epochs were completed. If the accuracy remained insufficient after 2000 epochs, a new network was initiated, such that all 20 networks had at least a performance of $0.98$. For each experimental condition -- covering both tasks and six different delay regimes -- 20 replicate training runs were executed.

\subsection{Information Relay Method and Recurrent States}
The concept of information relays is grounded in the notion of information channels. Each neural network layer comprises an input vector that is transformed through the computations of the layer's weights and potentially a threshold function, producing an output vector. As layers are sequentially stacked, the information flow from the input to the output is typically optimized through training. This information must be propagated through the network, layer by layer, ensuring its presence within the hidden nodes. The information relay method divides the nodes into two sets: the information-relaying nodes, denoted as $Y_R$, and the remaining nodes, denoted as $Y_0$. The information the set $Y_R$ relays can be quantified by Equation \ref{equ:relay}.

However, this does not reveal which specific nodes are relaying the information, only the amount of information relayed by a possible set $Y_R$ from inputs $X_{\rm in}$ to outputs $X_{\rm out}$. Testing all possible sets, of which there are $2^N$ where $N$ is the number of hidden nodes, would be computationally prohibitive as $N$ increases. Fortunately, a greedy algorithm can be employed to reduce this complexity. It starts with all nodes included in set $Y_R$ and $Y_0$ being empty. All possible nodes are tested by sequentially moving them to the non-relaying set $Y_0$ and identifying the set that retains the highest amount of relayed information. This identifies the node with the least contribution, which can then be permanently moved to $Y_0$. This process is repeated until only one node remains in $Y_R$. The result is a sequence of nodes, from the one relaying the least information to the one relaying the most information. We have previously shown that in strictly feed-forward neural networks (non-recurrent), the order of nodes also correlates with their functional contribution, as one would expect \cite{hintze2023detecting}.

Note that this process can be applied to subsets of information. Given that each task encapsulates three distinct concepts (rows one, two, and three in the memory task and size, direction, and brightness in the block task), the aforementioned process is repeated for each individual concept. This approach allows us to identify which nodes are relaying information and distinguish between specific aspects of the task.

The data extracted from determining the order of nodes and the amount of information they relay can be organized into a matrix that displays, for each node, the amount of information relayed for each aspect of the task. In practical terms, this translates to a matrix of $12 \times 3$.

While the above methodology has proven effective in non-recurrent neural networks, recurrent networks might operate differently, as the information that each recurrent node stores or processes could vary from one time point to another. Thus, we compute relay information for each possible time point independently and only after all the information has been input into the network. This approach ensures that we do not need to account for ongoing computations and, in principle, allows us to observe whether information migrates from one node to another over time. For instance, if at a hypothetical time point $t$, all information is held in nodes 1, 2, and 3, it could shift to nodes 4, 5, and 6 at time point $t+1$. By documenting the location of the information at each time point, we can track these changes.

\section{Results}
Initially, 20 independent replicates of RNNs, GRUs, and LSTMs were trained on each of the two time series classification tasks (``memory'' and ``block'') across six different modes of delay. Each delay mode incorporates additional empty vectors fed into the networks before the final output is obtained. This design allows networks to generalize across the temporal domain. It has been noted, particularly for RNNs, that they are highly sensitive to the precise length of a task sequence \cite{hafiz2022predicting}. For example, if an RNN is trained for a sequence of 10 time points, it may perform flawlessly when responding at the 11th time point but fails if the sequence extends to 11 times, with the last one containing no new information. Consequently, the networks were trained under specific delays of 1, 2, 3, 4, and 5 time points and under a condition where the delay was randomly selected from the interval $[1,5]$. It is anticipated that networks trained with a fixed delay will perform optimally within that specific interval and significantly worse outside of it, whereas networks trained with a random delay are expected to demonstrate better generalization across varying time points.

\begin{figure*}[htbp]
\centering
\includegraphics[width=0.85\textwidth]{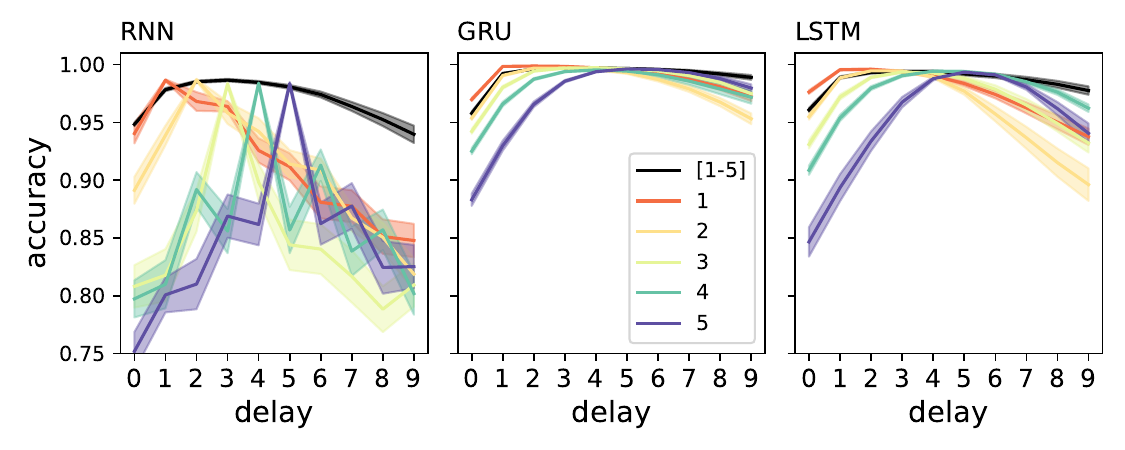}
\includegraphics[width=0.85\textwidth]{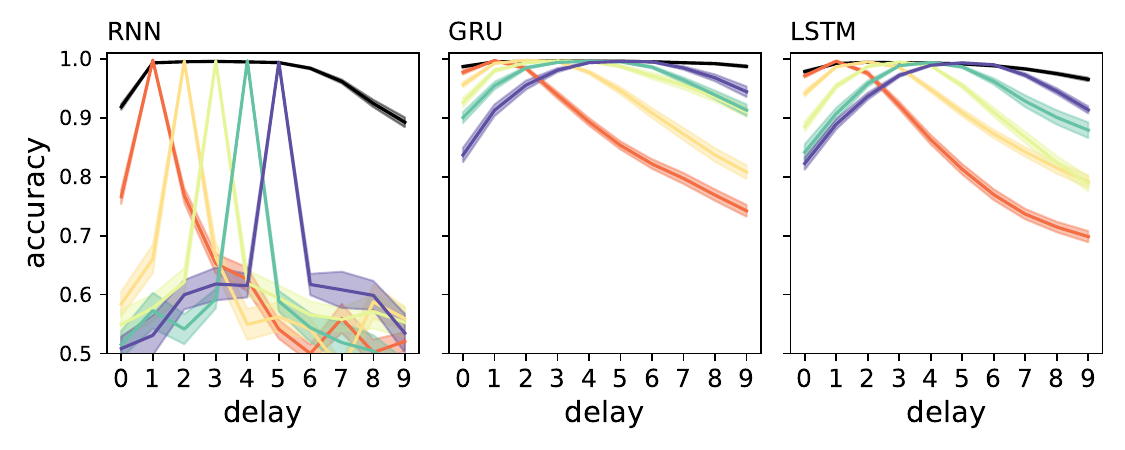}
\caption{Performance of different recurrent neural networks (RNN, GRU, and LSTM) independently trained for two different tasks. At the top row, results for the \textit{memory} task, and for the \textit{block} task at the bottom. Networks were trained with either a specific delay of 1, 2, 3, 4, or 5 time points between the information was delivered and the answer obtained (see color legend), or trained with randomly picking on of those 5 delay intervals for each input sequence. Performance was measured for 10 different delays ranging from 0 to 9 - observe that the time delays (for example, 0, 6, 7, 8, or 9) were never observed during training, and thus, performance measured for those delays characterizes the ability of the networks to generalize to the temporal domain. The shadow behind each line shows the standard error within all 20 experiments.}\label{fig:performance}
\end{figure*}

\begin{figure}
\centering
\includegraphics[width=0.475\textwidth]{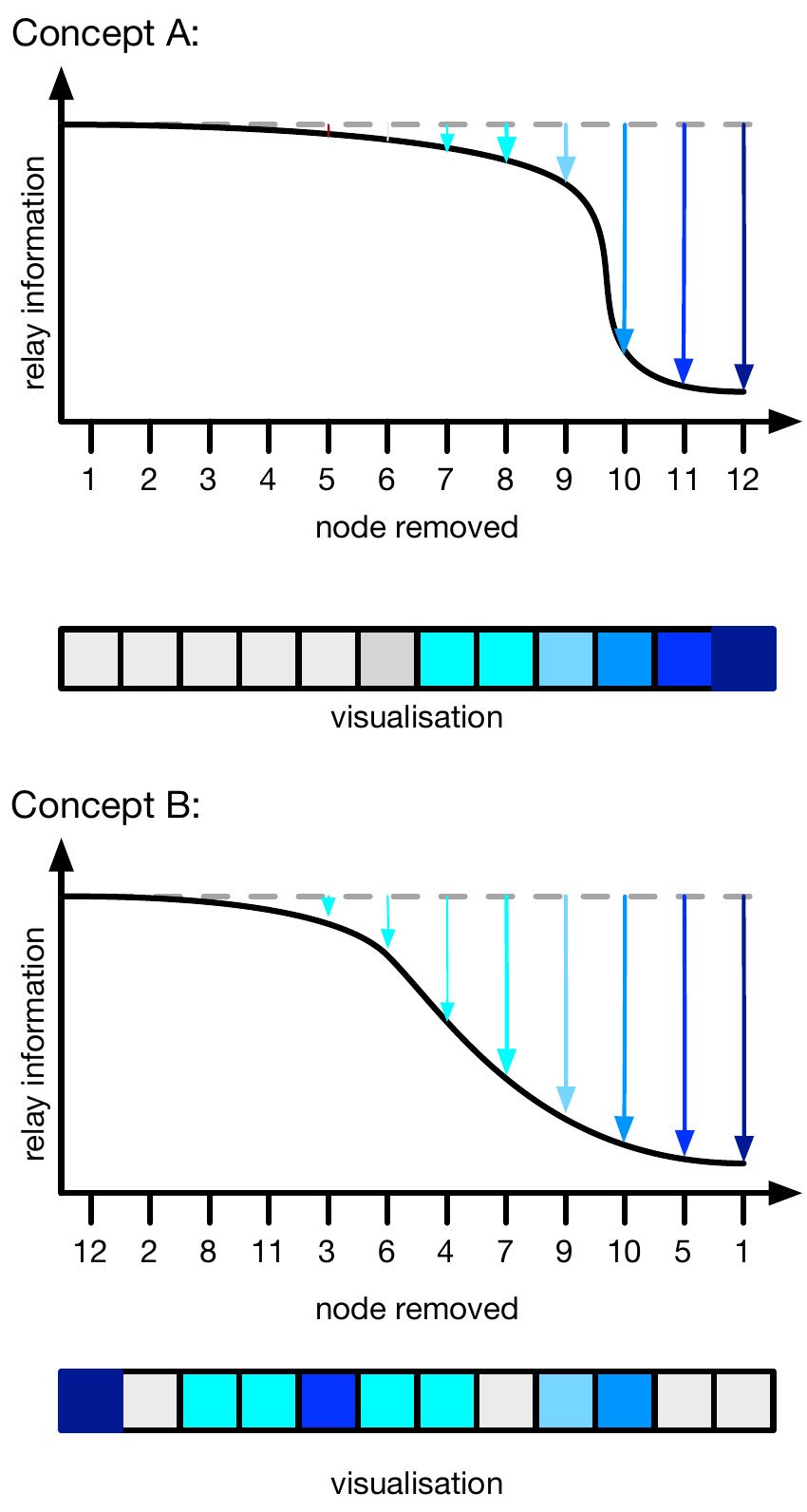}
\caption{Visualization of the data the information relay method and the greedy algorithm produces. On top the decrease of information about a hypothetical concept A the set of nodes relays while the set is continuously becoming smaller (black line). The reduction is shown as a plot, and the relay information loss as blue arrows. For simplicity reasons it is assumed here that the order in which nodes are removed goes from 1 to 12, as the later nodes relay the most information. Consequently, the amount of information each node relays can be visualized in the form of a bar at the bottom of the plot. Below that, the same process is shown for a different concept B, the nodes of a network can relay. Here the order of nodes is different, and thus the visualization highlights different nodes. }\label{fig:relayIllustration}
\end{figure}

The information relay method calculates the information transmitted from inputs to outputs through a set of relaying nodes distinct from the non-relaying nodes. To pinpoint the correct nodes, a greedy algorithm initiates with the entire set and progressively removes the node contributing the least. Consequently, one outcome is a sequence of nodes increasingly relaying more information (refer to Figure \ref{fig:relayIllustration}). It's important to note that as nodes are removed, the total relayed information is expected to decrease, yet the contribution, defined as the total information minus the actual relayed information, should increase (as depicted by increasingly longer arrows in Figure \ref{fig:relayIllustration}). The sequence in which nodes are removed often does not correspond to their order in the hidden layer. To illustrate the contribution of each node, the accumulated loss for each node is displayed (see Figure \ref{fig:relayIllustration} visualization).

In the context of non-recurrent neural networks, we employed noise to confirm that nodes identified by the information relay method are functional. The more essential a node is, the more disruption is caused when noise is introduced into that node, impacting the overall network function. However, dealing with recurrent neural networks presents different challenges, where future states depend on all previous nodes due to the fully connected structure of the recurrent layer. Consequently, introducing noise to a single node at a specific time point $t$ impacts all subsequent nodes at $t+x$. Moreover, we observed that information is relayed across multiple nodes, suggesting that applying noise to a single node might not significantly disrupt function, given that neural networks exhibit some robustness to noise applied to individual nodes.

In contrast, the cell states of an LSTM are not confined to a range of $[-1,1]$ as they are not limited by a hyperbolic tangent function, meaning noise impacts RNNs and LSTMs differently. Therefore, we adopt a different approach termed ``knockout,'' where sets of hidden nodes are set to $0.0$, analogous to observing the functional phenotype of a gene knockout.

Each task involves relaying three distinct concepts; for the memory task, these are three bits, and for the block task, these are direction, size, and brightness. For a trained network, we compute the sequence of information relaying sets using the greedy algorithm, starting with the entire set and progressing until only a single node remains. The relay information is computed at the time point after all information has been fed into the network, which is also when we perform the node knockouts. Initially, when all nodes are included and subsequently knocked out, we anticipate a complete loss of function. As the set size decreases, the effect of the knockout diminishes, and we expect a lesser loss of function. However, nodes identified as relaying specific concept information, such as concept A or block size, are predicted to exhibit a more pronounced loss in functionality for those specific concepts when knocked out compared to nodes associated with other concepts. Moreover, we do not anticipate these functional losses to be isolated since nodes frequently participate in relaying information for multiple concepts simultaneously.

\begin{figure}
\centering
\includegraphics[width=0.475\textwidth]{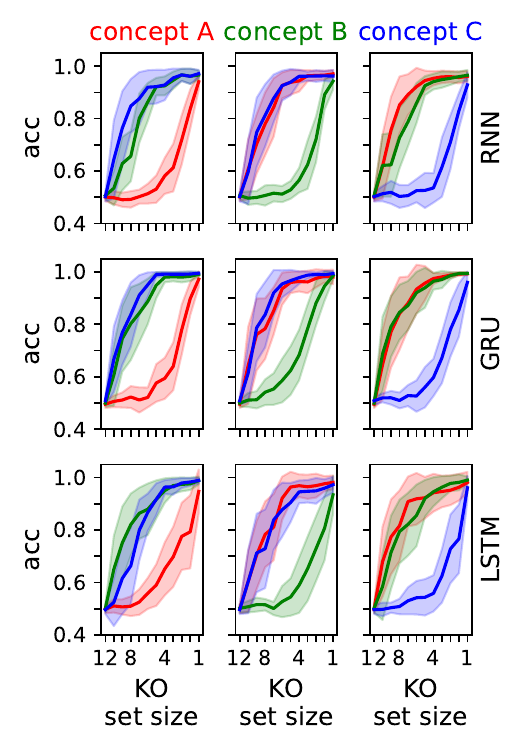}
\caption{Effect of knockouts (setting sets of hidden nodes to $0.0$ for different neural networks (RNN, GRN, and LSTM). The information relay method identified nodes carrying a specific concept (columns A, B, or C, see color). Knockouts were applied to the largest set first and then to progressively smaller sets in the order of importance to the given concept. Accuracy for each concept in shown in their respective colors. Results present averages over all 20 independently trained networks on the memory task. The shadows behind each line are 95\% confidence intervals of the mean. The delay used during training was from the interval $[1,5]$.}\label{fig:KOmemory}
\end{figure}

\begin{figure}
\centering
\includegraphics[width=0.475\textwidth]{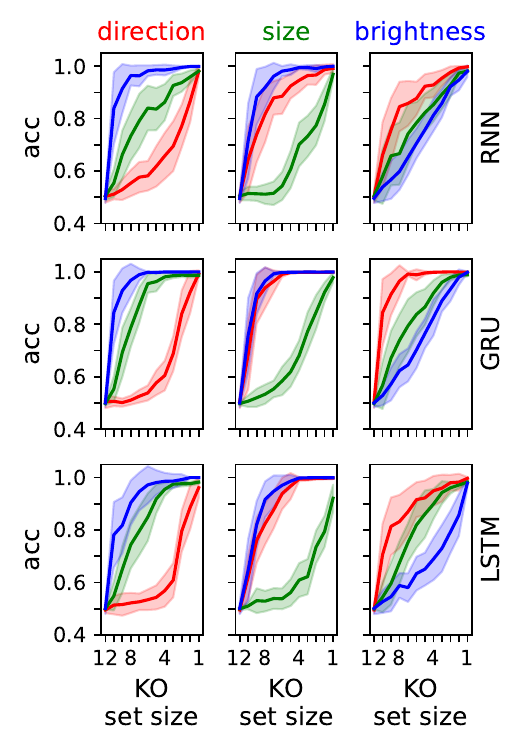}
\caption{Effect of knockouts (setting sets of hidden nodes to $0.0$ for different neural networks (RNN, GRN, and LSTM). Color code and layout is the same as in Figure \ref{fig:KOmemory}, except that here the block task was used, and thus the three concepts are direction, size, and brightness.}\label{fig:KOblock}
\end{figure}

Our findings confirm our expectations (refer to Figures \ref{fig:KOmemory} and \ref{fig:KOblock}) that knockouts applied to nodes identified to relay specific concepts indeed cause a more substantial loss in function for the classification of those concepts than for others. Conversely, knocking out nodes not identified as specific concept relays does not significantly affect function. Moreover, as expected, knocking out large sets of nodes generally disrupts function. Since all networks and both tasks consistently show this correlation between nodes identified to relay information and their knockout, leading to a loss of function for specific tasks, we conclude that the information relay method is also effective for recurrent neural networks. However, so far, we have demonstrated that the information relay method is effective at the specific time point when all information has been presented to the network. This might not pose a major issue if we assume that information stored in a set of hidden nodes remains within that specific set over multiple updates. Nevertheless, it is conceivable that information about a specific concept could transition from one set of nodes to another at each time point \cite{bohm2022understanding}. Therefore, we will now apply the information relay method to identify where the information is and when it is present. We expect that once information is acquired by the recurrent neural network, it should be retained in a specific set of nodes to facilitate retrieval at a later time. This is precisely how an LSTM is designed, setting and resetting cell states to preserve long-term memory. It would not be logical otherwise.

\begin{figure}
\centering
\includegraphics[width=0.475\textwidth]{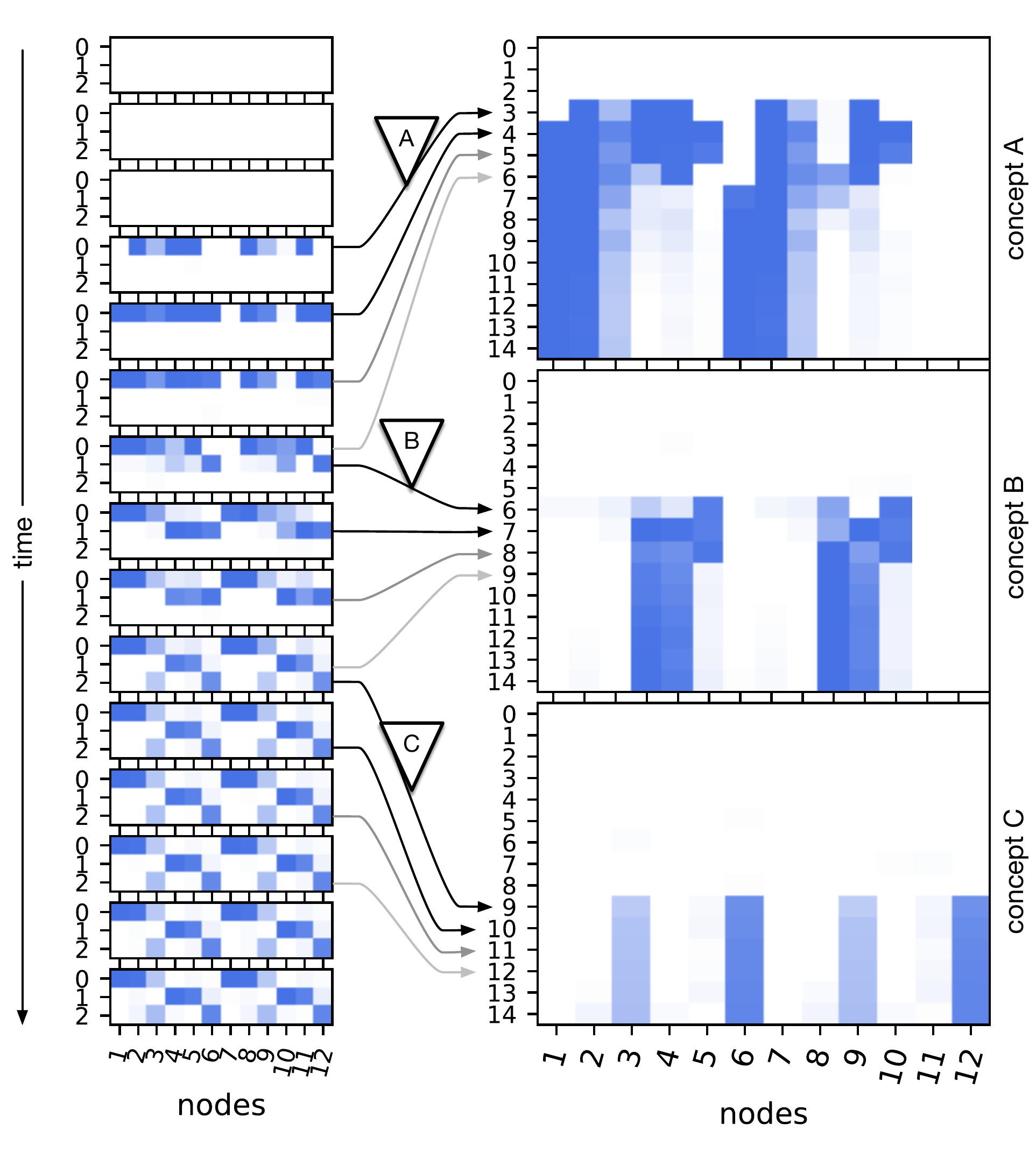}
\caption{Illustration of information that each node carries about each concept identifed by the information relay method. On the left, for the memory task, where information about the three concepts A, B, and C are given at time points 4, 7, and 10 (see triangles) the sequence of I matrices. Each of the 15 matrices on the left, shows in shades of blue the amount each node (12 columns) stores about each concept (3 rows). On the right, each panel shows the information of each of the 12 nodes about one of the concepts (see right hand label for concept) over the progression of 15 time points. The arrows illustrate how the right panels were constructed from the succession of I matrices. The network show here is a perfectly trained LSTM.}\label{fig:timeSeriesIllustration}
\end{figure}

To demonstrate that the information relay method can identify information at different time points, we utilize the memory task, injecting concept A precisely at time point 4, concept B at time point 7, and concept C at time point 10. As expected (refer to Figure \ref{fig:timeSeriesIllustration}), our example task applied to an LSTM shows that the information relay method not only identifies when the information arrives at a particular node but also observes that, following the initial updates, the information generally remains in the same nodes throughout the process in the case of an LSTM. Additionally, we notice that some nodes appear to store information about more than one concept—a phenomenon of information ``smearing'' that we have observed in many other contexts previously. This finding further validates the capability of the information relay method to track and analyze the dynamic handling of information within recurrent neural networks, specifically LSTMs, over time.

\begin{figure}
\centering
\includegraphics[width=0.475\textwidth]{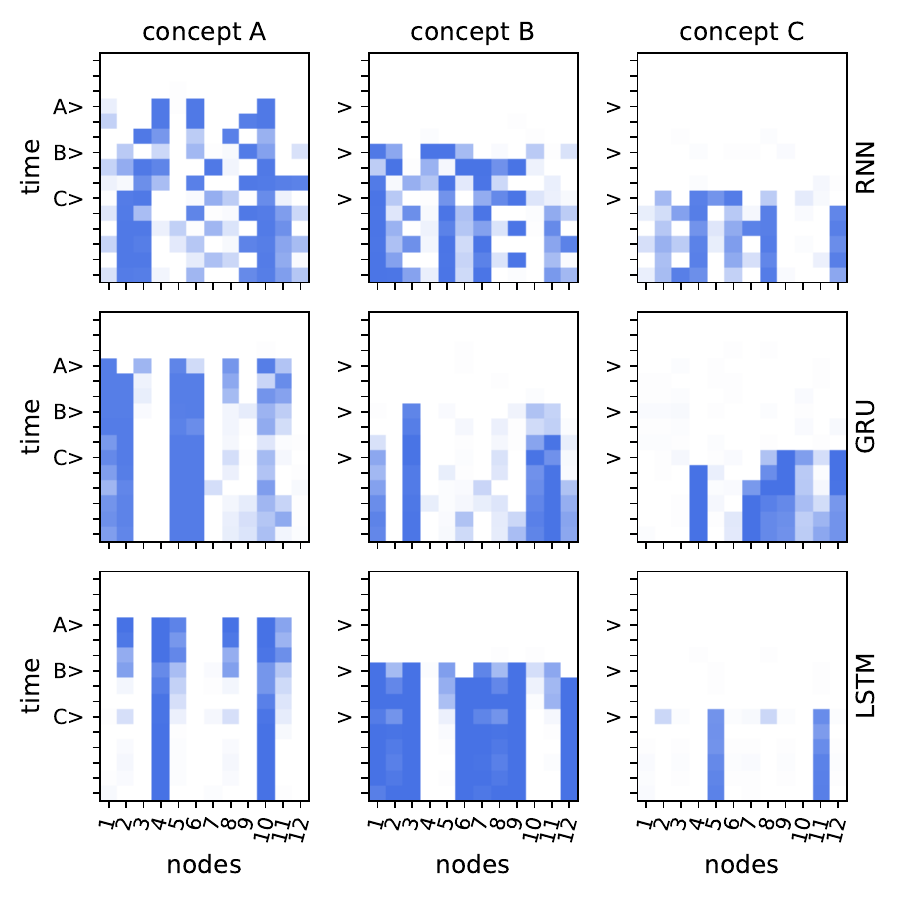}
\caption{Information distribution for all 12 nodes, given the three concepts (A, B, and C) from the memory task, for an RNN, GRU, and LSTM (see row label on the right-hand side). Information content is shown in blue shades, with the darkest blue corresponding to 1 bit of information. On the right, labeled with A, B, and C are the time points at which the information was presented to the networks.}\label{fig:compareLocalization}
\end{figure}

When comparing the three different types of recurrent models (RNN, GRU, and LSTM) on the same task, where information is presented at three distinct time points (4, 7, and 10), notable differences in how information is localized become apparent (refer to Figure \ref{fig:compareLocalization}). In the cases of the GRU and LSTM, information is absorbed into the hidden or cell states and, following a brief equilibration phase of about three time points, remains localized within those states. Conversely, in the RNN, the information does not stay localized but instead \textit{moves around}; that is, the location of information changes from one time point to the next. This dynamic illustrates a fundamental distinction in how these models handle temporal data, highlighting the RNN's tendency to redistribute information across its network over time, unlike the GRU and LSTM, which stabilize and retain information more consistently in specific nodes.

\begin{figure}
\centering
\includegraphics[width=0.475\textwidth]{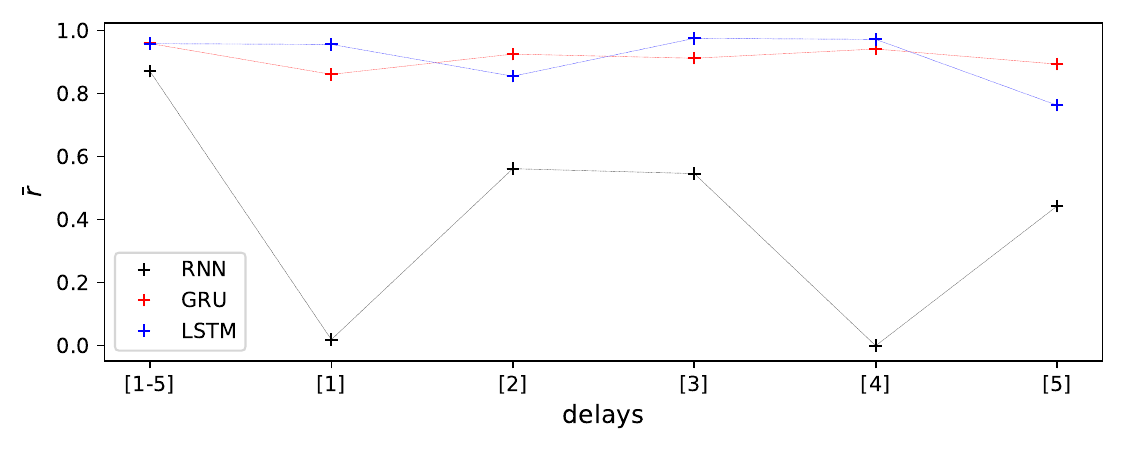}
\includegraphics[width=0.475\textwidth]{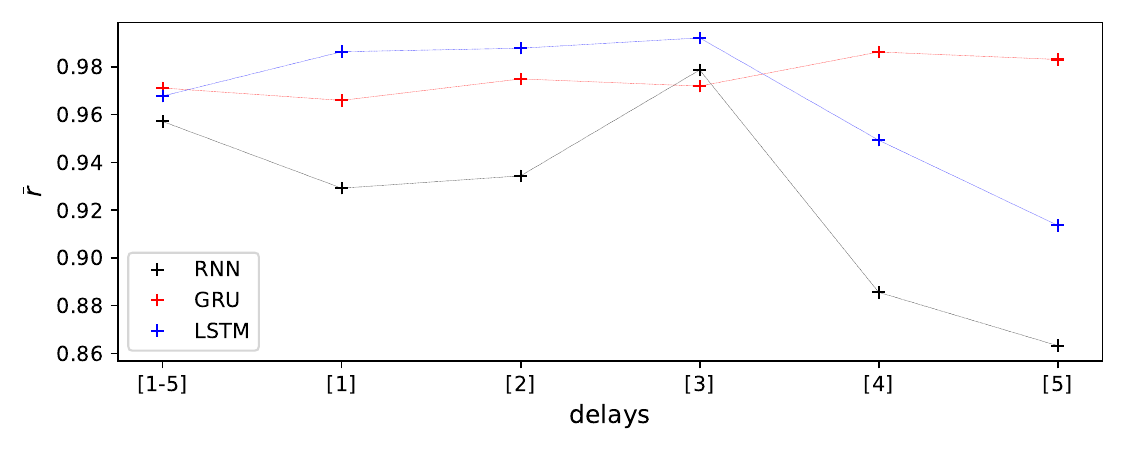}
\caption{Correlation coefficient $r$ between nodes measured at different time points shown for three different types of neural networks (RNN in black, GRU in red, and LSTMs in blue) averaged over 20 independent replicates for the memory task (top panel) and the block task (bottom panel). The x-axis shows the different kinds of delays experienced during training.}\label{fig:ccBoth}
\end{figure}

\begin{figure}
\centering
\includegraphics[width=0.475\textwidth]{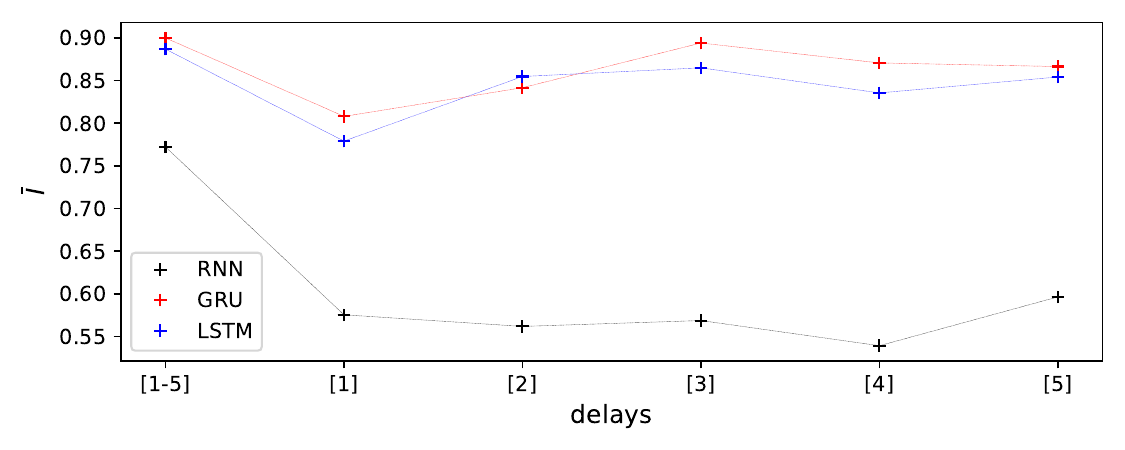}
\includegraphics[width=0.475\textwidth]{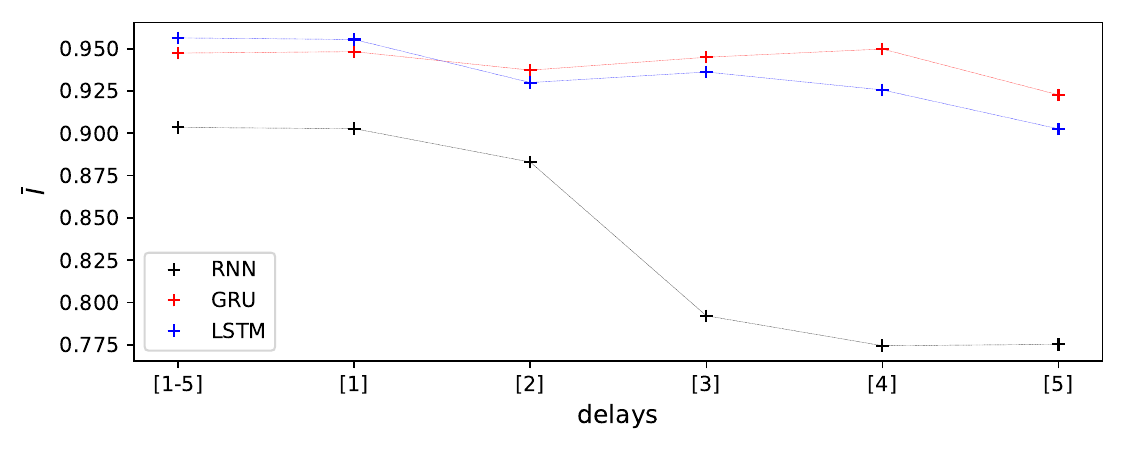}
\caption{Information overlap $I$ between nodes measured at different time points shown for three different types of neural networks (RNN in black, GRU in red, and LSTMs in blue) averaged over 20 independent replicates for the memory task (top panel) and the block task (bottom panel). The x-axis shows the different kinds of delays experienced during training.}\label{fig:iBoth}
\end{figure}

The tendency of RNNs to shift information between nodes appears to be a characteristic feature of their architecture. In both tasks, we observe a higher correlation in the distribution of information between time points in GRUs and LSTMs compared to RNNs, as illustrated in Figure \ref{fig:ccBoth}. This suggests that GRUs and LSTMs maintain a more stable distribution of information over time. Additionally, the overlap of nodes carrying the same information is consistently lower in RNNs than in GRUs and LSTMs, as shown in Figure \ref{fig:iBoth}. This indicates less consistency in node function across different instances within RNNs.

Interestingly, when networks are trained to respond to a range of delays ([1-5]), RNNs exhibit a tendency to localize information similarly to GRUs and LSTMs. However, this localization pattern shifts dramatically when the networks are trained to respond at a specific time point ([1] to [5]). Under these conditions, the characteristic phenomenon of information movement within RNNs becomes evident. This suggests that the training context—specifically, the variability in delay—plays a crucial role in how information is handled and stabilized within the network architecture.

We have demonstrated that the information relay method can accurately identify which nodes carry information, confirming that this information is functionally relevant. Furthermore, the method can be applied at various time points, revealing that in RNNs, information dynamically moves from node to node over time—more so than in GRUs or LSTMs. An interesting aspect arises concerning the overlap of information within specific nodes. It seems that individual nodes may relay information about multiple concepts, not just a single one. This process could be conceptualized through the idea of encoding or encryption. Consider a node that carries information about two different attributes, such as the size (small or large) and brightness (bright or dark) of a block. The node could exhibit four distinct states—e.g., $-1.0$, $-0.2$, $0.2$, and $1.0$. Here, negative values ($-1.0$ and $-0.2$) could signify a dark object, whereas positive values indicate a bright one. Simultaneously, the smaller absolute values ($\pm0.2$) could denote a small object, and the larger absolute values ($\pm1.0$) a large one. This method of encoding allows a single node to represent multiple concepts simultaneously, though this can be far more complex than illustrated here.

\begin{figure}
\centering
\includegraphics[width=0.475\textwidth]{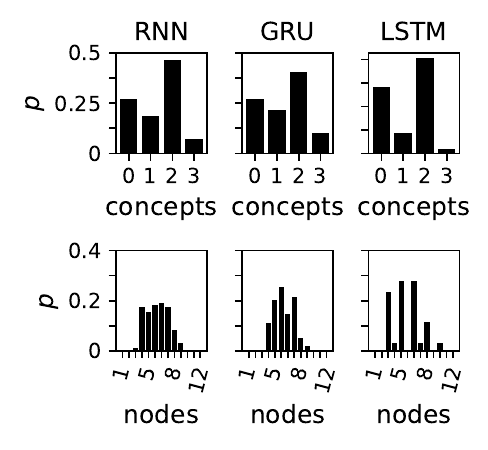}
\caption{Fraction of nodes relaying differently many concepts (top panel) or the number of nodes carrying the same concept (bottom panel). Averages were obtained from 20 independently trained networks (see title for type) on the memory task.  }\label{fig:usageMemory}
\end{figure}

\begin{figure}
\centering
\includegraphics[width=0.475\textwidth]{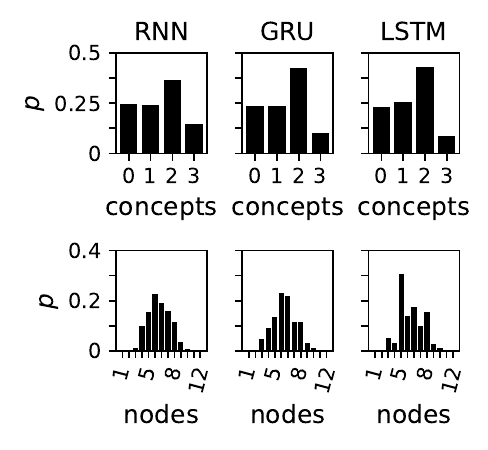}
\caption{Fraction of nodes relaying differently many concepts (top panel) or the number of nodes carrying the same concept (bottom panel). Averages were obtained from 20 independently trained networks (see title for type) on the block task.}\label{fig:usageBlock}
\end{figure}

To further explore this, we examine the matrix output from the information relay method (refer to Figure \ref{fig:timeSeriesIllustration}, left column). This matrix, consisting of 3 rows (one for each concept) and 12 columns (one for each node), shows the amount of information each node relays about each concept. Using k-means clustering with $k=2$, we convert this matrix into a binary format that indicates which nodes are involved in relaying which concepts. By summing the values across each row and column, we assess how many nodes are involved in relaying a particular concept and how many concepts each node relays, respectively. Histograms of these normalized sums (see Figures \ref{fig:usageBlock} for the block task and \ref{fig:usageMemory} for the memory task) reveal that individual concepts are not confined to single nodes but are instead distributed across multiple nodes, irrespective of the network type or task at hand. This finding underscores the complexity and distributed nature of information processing within neural networks.

The analysis presented above demonstrates the effectiveness of the information relay method for identifying how information is stored and transitions across multiple recurrent updates. However, this approach diverges from the previously discussed analysis of the latent space of the hidden variables. In that method, principal component analysis (PCA) is applied to the hidden state variables to achieve a lower-dimensional projection (typically 2D), where the different states of the neural network often form clusters or reveal further insights into the decision-making space of the trained model. This method applies equally to the tasks and networks addressed in this study.

\begin{figure}
    \centering
    \includegraphics{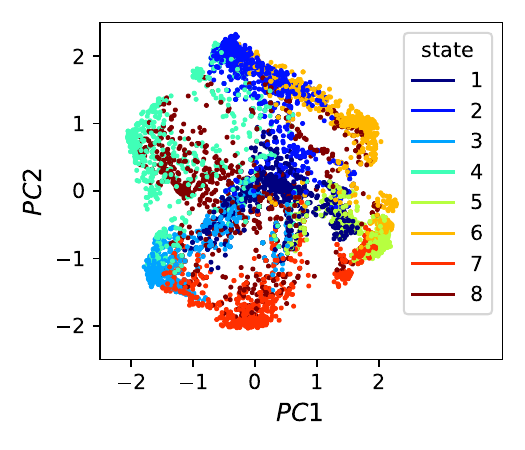}
    \caption{Principle Component Analysis for the hidden states of an RNN trained on the memory task. Colors indicate the 8 possible states the network can be in as it remembers the 3 combinations of binary variables.}
    \label{fig:regularPCA}
\end{figure}

We conducted a PCA for a single RNN trained on the memory task. Here, the hidden states are represented by a 12-dimensional vector, which is projected into a 2D space to maximize the variance between individual points. Given that the network is designed to retain three independent binary values, it can theoretically exist in up to 8 different states. The projected hidden states in this PCA are color-coded according to the state of the network, defined by the three variables it is tasked to remember. The latent space effectively resolved these 8 states, suggesting that the trained network likely forms one attractor for each potential state, as illustrated in Figure \ref{fig:regularPCA}. This method not only supports the visualization of network processing dynamics but also complements the findings from the information relay method by providing a spatial representation of how the network organizes and retains information across different tasks.

While the analysis of the latent space provides valuable insights into how it is organized around specific tasks, it lacks the capability to track information flow and, crucially, it does not assess the importance of individual nodes. In contrast, the information relay method not only identifies the significance of each node but also proposes a sequence for node removal from the PCA, enhancing interpretability.

\begin{figure*}[htbp]
    \centering
    \includegraphics[width=0.95\textwidth]{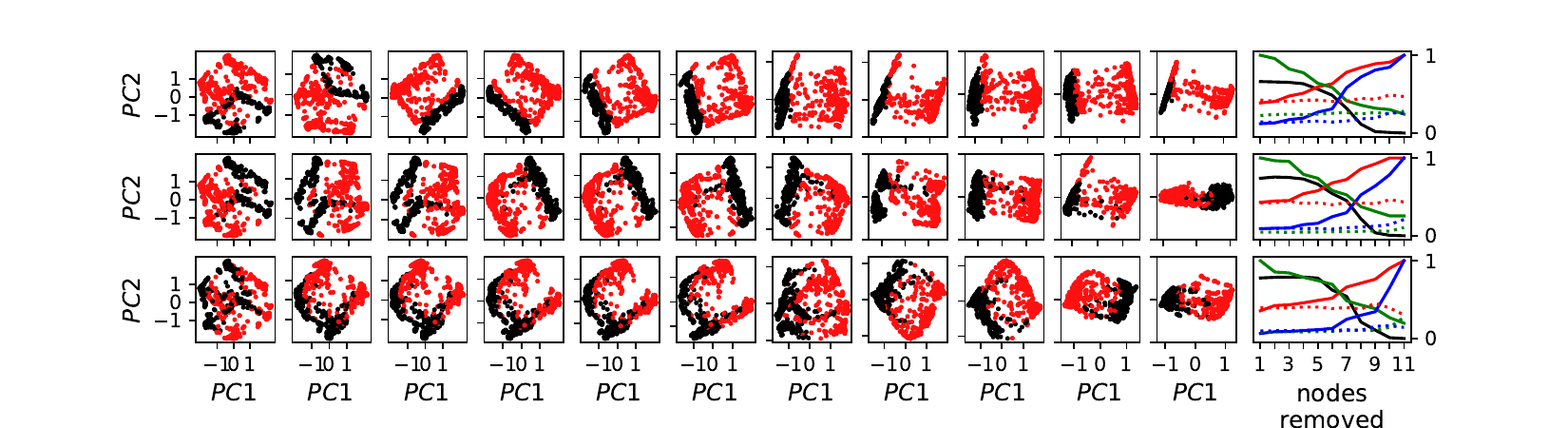}
    \caption{Principle Component Analysis based on the nodes identified by the Information Relay method. Panels on the left show a PCA performed on an RNN trained on the memory task; however, an increasing number of hidden nodes was removed from left to right, identified by the information relay method (information shown as a black line in the right-most panel). Each row represents one of the three variables to be remembered by the RNN. Individual points of the PCA are color-coded by the state they represent. In black is the state of the variable in question for each row, and in red is the remaining one. The colored lines in the most right panel show how well the red and black nodes cluster. In red, the silhouette score (the larger, the better clusters separate); in blue, the Davies Bouldin Score (which indicates better separation also with a higher score); and in green, the Calinski Harabasz Score (which becomes smaller, the better clusters separate). As dotted lines, the same clustering methods were applied to the PCA, where nodes were removed in a random order, using 50 replicated samples and computing averages.}
    \label{fig:PCAanalysis}
\end{figure*}

If nodes are removed in a random order, we anticipate minimal or even negative impacts on the PCA, potentially obfuscating the formation of meaningful clusters in the latent space. However, systematically removing nodes starting with the least important in terms of information relay can refine the PCA, making the latent space more informative as less relevant nodes are excluded. This approach helps in isolating the core structural elements that are critical for the network's information processing capabilities. We can observe the effectiveness of this targeted removal strategy in Figure \ref{fig:PCAanalysis}, where removing less critical nodes progressively sharpens the focus on the clusters that define meaningful attractors, thereby providing deeper insights into the neural network's operational dynamics.

This integration of methods illustrates several key points. Initially, the improved accuracy in the separation of clusters as nodes are selectively removed --contrasting with the effects of random removal -- reinforces the validity of the information relay method. This outcome not only supports the method's effectiveness but also demonstrates that these techniques can be synergistically combined. Importantly, the deliberate removal of irrelevant nodes from the latent space enhances the precision of the PCA. This process clarifies the underlying structure of the data by focusing the analysis on nodes that contribute significantly to the network's function, thereby providing more insightful and accurate visual representations of the neural network's operational dynamics.

\section{Discussion}
The study provides several critical insights into the functioning of artificial and recurrent neural networks, advancing our understanding beyond traditional performance metrics to the detailed roles of individual nodes and their dynamics over time. Through the application of the information relay method, we have demonstrated that neural networks are not mere black boxes; rather, they have discernible internal structures and patterns of information flow that can be systematically analyzed and comprehended.

By evaluating RNNs, GRUs, and LSTMs on two distinct tasks that required the networks to select and store information from inputs across several time points, we confirmed that the information relay method is effective not only for non-recurrent systems but also in these more complex scenarios. Additionally, by monitoring information flow over time, we have identified unique patterns in how each model manages its internal states. GRUs and LSTMs exhibit clear localization of representations, with information remaining concentrated within specific sets of nodes. In contrast, RNNs demonstrate a drift in their information storage, with representations shifting between nodes at each update. This behavior suggests that RNNs face challenges in preserving information over extended sequences, likely contributing to their comparatively weaker performance on tasks involving longer time lags relative to GRU and LSTM variants. These findings highlight the superior capabilities of GRUs and LSTMs in managing temporal information. 

The observation that information in RNNs tends to diffuse randomly raises important considerations for the dynamics of recurrent networks, suggesting areas for further research into more complex time-series challenges. This insight could be pivotal in developing strategies to enhance the stability and efficiency of RNNs for a broader range of applications.

This method has several limitations. First, it was demonstrated only on two simple time series classification tasks, and more complex tasks might impose additional challenges to training and testing representations. However, we do not foresee any significant limitations beyond a possible scaling to extremely large networks. We also did not perform a hyperparameter search; however, we believe this would not fundamentally alter our findings, as no single hyperparameter is likely to counteract the effects of representation smearing observed completely. Additionally, the simplicity of the tasks was intentional to clearly illustrate how the Information Relay method operates with a minimal set of variables. Lastly, we did not incorporate dropout in our experiments. This decision was based on existing research \cite{sardar2023robustness}, which indicated that dropout tends to increase functional smearing, although it was not specifically investigated in a recurrent context.



 
\bibliography{sample.bib}

\bibliographystyle{IEEEtran}




\end{document}